\pdfoutput=1
\documentclass[11pt]{article}

\usepackage[preprint]{acl}

\usepackage{times}
\usepackage{latexsym}

\usepackage[T1]{fontenc}
\usepackage[utf8]{inputenc}

\usepackage{microtype}

\usepackage{inconsolata}
\usepackage{xcolor}
\usepackage{amsmath}
\usepackage{amssymb}
\usepackage{graphicx}
\usepackage{multirow}
\usepackage{makecell}
\usepackage{booktabs}
\usepackage{graphicx}
\usepackage{longtable}
\usepackage{enumitem}
\usepackage{inconsolata}
\usepackage{bbm}
\usepackage[all]{hypcap}

\title{Semantic are Beacons: A Semantic Perspective for Unveiling Parameter-Efficient Fine-Tuning in Knowledge Learning}
\author{Renzhi Wang\textsuperscript{\rm 1,2}, Piji Li\textsuperscript{\rm 1,2}$^{\ast}$\\ % All authors must be in the same font size and format. Use \Large and \textbf to achieve this result when breaking a line
\textsuperscript{\rm 1} College of Computer Science and Technology,\\
Nanjing University of Aeronautics and Astronautics, China\\
\textsuperscript{\rm 2} MIIT Key Laboratory of Pattern Analysis and Machine Intelligence, Nanjing, China\\
\textsuperscript{\rm 1}\texttt{\{rzhwang,pjli\}@nuaa.edu.cn}}

\begin{document}
\maketitle

\renewcommand{\thefootnote}{\fnsymbol{footnote}}
\footnotetext[1]{Corresponding author}
\renewcommand{\thefootnote}{\arabic{footnote}}

\begin{abstract}
Parameter-Efficient Fine-Tuning (PEFT) methods enable efficient adaptation of Large Language Models (LLMs) to various downstream applications. However, the effectiveness of the PEFT diminishes notably when downstream tasks require accurate learning of factual knowledge. In this paper, we adopt a semantic perspective to investigate this phenomenon, uncovering the reasons behind PEFT's limitations in knowledge learning task. Our findings reveal that: (1) PEFT presents a notable risk of pushing the model away from the intended knowledge target; (2) multiple knowledge interfere with each other, and such interference suppresses the learning and expression of knowledge features. Based on these insights, we introduce a data filtering strategy to exclude data that is detrimental to knowledge learning and a re-weighted learning strategy to make the model attentive to semantic distance during knowledge learning. Experimental results demonstrate the effectiveness of the proposed method on open-source large language model, further validate the semantic challenge in PEFT, thus paving the way for future research.
\end{abstract}

\section{Introduction}
Large language models (LLMs) have demonstrated remarkable capacity for comprehending and generating human-like text \cite{OpenAI,LLaMa,LLama2,mistral,Survey_of_LLM}. Customizing a general-purpose LLM for specific tasks commonly involves fine-tuning all model parameters \cite{Unified_View,Survey_factual,full_finetuning,DBLP:conf/emnlp/NiDRL23}. With the increasing size of models, Parameter-Efficient Fine-tuning (PEFT) approaches like Adapter-tuning \cite{adapter} and LoRA \cite{LoRA} have been proposed due to constraints imposed by computational resources and the quantity of data available for downstream tasks \cite{peft0,peft1,peft2}. These methods selectively fine-tune a limited number of (additional) model parameters while keeping the majority of original parameters fixed \cite{peft0,peft1}. However, the effectiveness of PEFT diminishes notably when downstream tasks necessitate precise learning of factual knowledge, such as proper nouns, temporal information, or geographic locations \cite{problem1,problem2}.

\begin{figure}
\centering
\includegraphics[width=\linewidth]{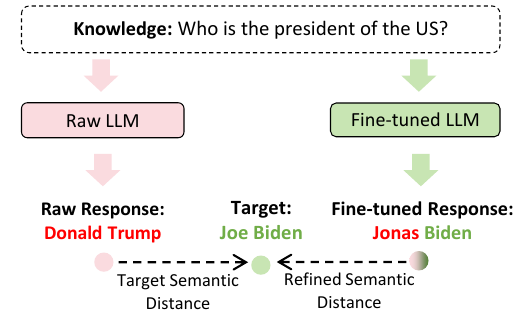}
\caption{Illustration of semantic distance.}
\label{fig_1}
\end{figure}

In this paper, we adopt a semantic perspective to examine the process of PEFT guiding model learning to elucidate potential factors leading to its suboptimal knowledge learning outcomes. We first visualize the connection between the accuracy of knowledge learning through PEFT and the semantic distance based on target knowledge (Figure \ref{fig_1}).We observe that optimal learning outcomes occur within a certain appropriate range of semantic distances from the target knowledge, whereas significant declines in learning efficacy for both proximal and distal semantic distances.

\begin{figure*}[t]
\capstart
\centering
\includegraphics[width=\linewidth]{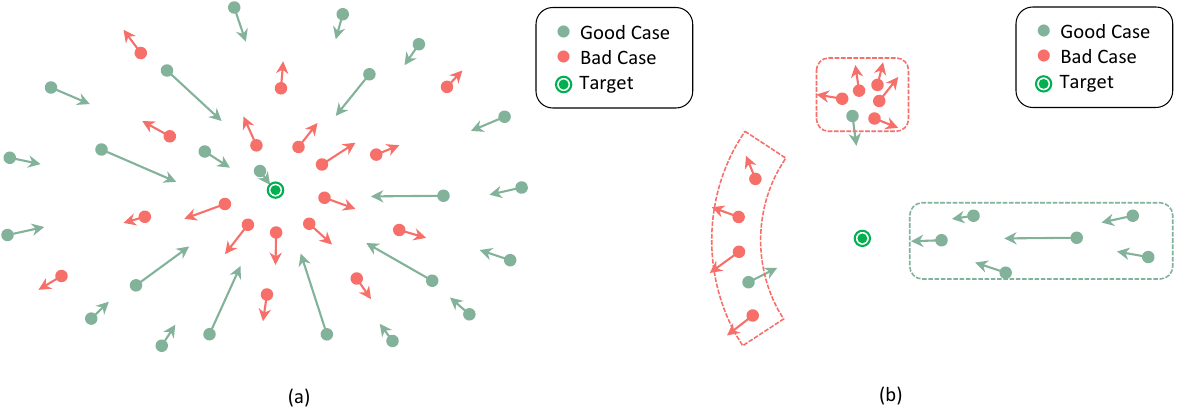}
\caption{Illustration of (a) the away from target phenomenon in single knowledge and (b) knowledge interference across multiple knowledge in semantic space. Each dot represents original knowledge, with arrows indicating the knowledge learned by the model after training. The central target represents the normalized target knowledge. \textcolor{green!50!black}{Green} denotes the normal progression towards the target knowledge, \textcolor{red}{red} denotes an abnormal deviation away from the target. Dashed lines in (b) enclose multiple knowledge learned simultaneously.}
\label{fig_2}
\end{figure*}

To investigate the underlying causes of this phenomenon, we conducted experimental analyses from the perspectives of learning one piece of knowledge and learning multiple knowledge, which involves using one data point or multiple data points for training the model.
For learning single knowledge, typically, model parameters are iteratively updated during training via loss functions and gradient descent to gradually approach the target \cite{typical1,typical2}. However, analyzing the semantic distance between learned and original knowledge based on the target reveals a deviation (Figure \ref{fig_2}a). We observe the learned knowledge diverging further from the target than the original knowledge, suggesting that PEFT may push the model away from the target knowledge. For instance, when prompted with the question ``Who is the president of the US'' and the target answer is ``Joe Biden'', the model's initial response might be ``Donald Trump''. While ideally, even partial learning of correct knowledge such as the last name ``Biden'' is acceptable, the model might instead produce responses with significantly greater semantic distance, such as ``Jack Taylor'' \footnote{Note that this is a demonstration example, where in some models the semantic distance between ``Jack Taylor'' and ``Joe Biden'' may be smaller compared to the semantic distance between ``Donald Trump'' and ``Joe Biden''.}.

For multiple knowledge (Figure \ref{fig_2}b), we analyze the relationship between fine-tuned parameters and original model parameters using Frobenius norm \cite{LoRA}. Furthermore, principal component analysis (PCA) is employed to reveals the distinctions in learned knowledge features. Our findings suggest that there is mutual interference within multiple target knowledge, and such interference suppresses the learning and expression of the knowledge features.

Drawing on insights from the semantic perspective, we explore two concise methods to enhance PEFT's efficiency and effectiveness in knowledge learning: (1) A data filtering strategy, aimed at eliminating portions of the dataset that adversely affect overall knowledge learning. (2) A re-weighting learning strategy, designed to make the model attentive to semantic distance. Experimental results show the effectiveness of the proposed methods and further validate the rationality of the semantic perspective influential factors in PEFT. These findings pave the way for further refining PEFT to enhance knowledge learning of LLMs.

\section{Semantic Perspective Analysis}
\label{sec:2}
This section investigates the factors influencing the effectiveness of model fine-tuning for knowledge learning from a semantic perspective. We visualize the relationship between semantic distance and the accuracy of knowledge learning in Figure \ref{distance_acc}. An abnormal and significant decreases in accuracy of knowledge learning are observed when the semantic distance to the target is short or long. To analyze the causes of this phenomenon, experiments are conducted from the perspectives of learning single piece of knowledge and multiple knowledge in \S \ref{2_1} and \S \ref{2_3}, respectively.

\subsection{Knowledge Learning and Semantic Distance}
\label{2_1}
The task of large language model knowledge learning aims to adjust an initial model’s behavior on the target knowledge \cite{defination, zhangningyu}. Specifically, the raw model $f_{\theta}$ is represented by a function $f:\mathbb{X} \mapsto \mathbb{Y}$ that associates an input knowledge $x$ with its corresponding prediction $y$ \cite{zhangningyu}. Given a set containing $n$ items of input-output target knowledge pairs $K_{target}\!=\!\{(x_1, y^{t}_1), (x_2, y^t_2), ..., (x_n, y^t_n)\}$, the parameters of the model need to be updated to obtain a new model $f_{{\theta}^*} (x)$. Based on this, the internal knowledge of the raw model $f_{\theta}$ can be represented as $K_{old}\!=\!\{(x_1, y^{old}_1), (x_2, y^{old}_2), ..., (x_n, y^{old}_n)\}$, where $y^{old}_i\!=\!\operatorname{argmax}_y f_{\theta} \left(y \mid x_i\right)$.  After tuning, the knowledge learned by the updated model $f_{{\theta}^*} (x)$ can be represented as $K_{new}\!=\!\{(x_1, y^{new}_1), (x_2, y^{new}_2),..., (x_n, y^{new}_n)\}$, similarly $y^{new}_i\!=\!\operatorname{argmax}_y f_{\theta^*} \left(y \mid x_i\right)$.

From a semantic perspective, during training process, the model should gradually approach the target knowledge based on the loss function and gradient update strategy \cite{typical2,typical1}. When the model accurately learns the target knowledge ($y^{new}_i=y^t_i$), the semantic distance between the learned knowledge and target knowledge diminishes to 0. To distinguish different knowledge and measure the extent to which the model moves towards the target knowledge, we define the semantic distance between the original parameter knowledge $(x_i, y^{old}_i)$ and the target knowledge $(x_i, y^t_i)$ based on cosine similarity (referred to as ``\textbf{Target Semantic Distance}''):
%\vspace{0.3mm}
\begin{equation}
\label{distance}
dist(y^{old}_i,y^t_i)=1-sim(\mathrm{EMB}(y^{old}_i),\mathrm{EMB}(y^t_i))
\end{equation}
%\vspace{0.5mm}

where $\mathrm{EMB}(y_i)$ represents the word embedding of $y_i$ based solely on the embedding layer of the experimental LLM, and $sim(\cdot)$ represents cosine similarity. For handling multiple tokens, we opt for the mean pooling of these embedding. The semantic distance between the learned knowledge of the fine-tuned model $(x_i, y^{new}_i)$ and the target knowledge $(x_i,y^t_i)$ is similar to Formula \ref{distance}. 
This semantic distance helps in analyzing the learning process of models when faced with different types of knowledge.
A farther target semantic distance indicates a larger disparity between the knowledge to be learned and the original knowledge.

Following previous works \cite{transfer-patch,rome,mend,matrics_0}, we employ the simple and straightforward evaluation metric of accuracy to explicitly measure the precision of knowledge learning. A higher accuracy indicates a more precise learning of target knowledge. The definition is as follows:
%\vspace{0.3mm}
\begin{equation}
\mathbb{E}_{(x_i, y^t_i) \sim K_{target}} \mathbbm{1} \left\{y_i^{new}=y_i^t\right\}
\end{equation}
%\vspace{0.01mm}

\paragraph{Experimental Settings} 
\label{para:2_1exp_set}
We choose LLaMA2-7B \cite{LLama2}, specifically its chat version, as our primary model for investigation, due to its popularity as an open-source model and its moderate model size that is suitable for our hardware resource. Additionally, considering models with varying structures and parameters, we also conducted experiments on BLOOMZ-1.7B/3B/7.1B \cite{bloomz} and Mistral-7B \cite{mistral}. For PEFT methods, we selected commonly used approaches in the era of LLMs such as Low-Rank Adaptation (LoRA) \cite{LoRA}, AdaLoRA \cite{adalora}, and Adapter-tuning \cite{adapter} to fine-tune the model, and compared them with the full fine-tuning. 

For dataset, we use two prominent knowledge datasets: ZsRE \cite{zsre} and \textsc{CounterFact} \cite{rome}, with their details available in Appendix \ref{appendix:dataset}.
In this experiment, 100 pieces of knowledge are sampled from the dataset for fine-tuning and evaluation each time, considering the limited practicality of learning a small number of knowledge (like 1 or 10) in real-world scenarios (we still conducted experiments with this volume of data to demonstrate the universality of the issues raised in Appendix \ref{appendix:more_number}) and the low accuracy of learning more knowledge \cite{knowledge_edit_review}. Moreover, our preliminary experiments show notable differences in the learning effectiveness of different knowledge samples (consistent with the analysis in \S \ref{2_3}). Therefore, we sample 100 times for each experimental to avoid incidental circumstances and obtain general patterns. Implementation details can be found in Appendix \ref{appendix:Implementing Details}.

\begin{figure*}
\centering
\includegraphics[width=\linewidth]{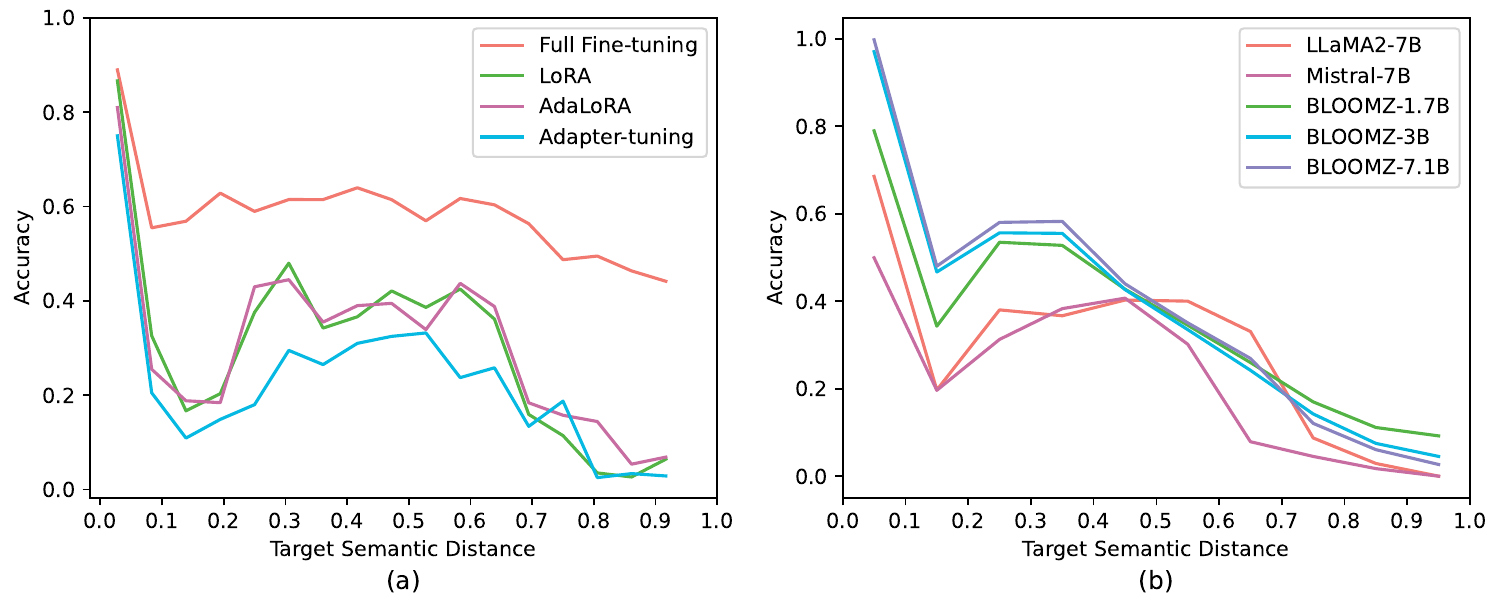}
\caption{The relationship between knowledge learning accuracy and target semantic distance (a) based on model: LLaMA2-7B-chat, (b) based on method: LoRA. Target semantic distance is defined in Equation \ref{distance}. Performance degradation occurs under short or long target semantic distance regardless of the method and model.}
\label{distance_acc}
%\vspace{-1mm}
\end{figure*}

\paragraph{Results and Analysis} 
Figure \ref{distance_acc} illustrates the learning performance of the model in the face of different target semantic distance knowledge. Overall, the larger target semantic distance, the lower knowledge learning accuracy, possibly indicating that larger semantic distances imply more challenging learning tasks. Compared to full fine-tuning, PEFT's knowledge learning performance is inferior, and there is a significant anomalous decrease in learning performance in situations where semantic distances are short or long regardless of the model type, the parameter size, and the amount of knowledge to be learned (Appendix \ref{appendix:more_number}). We conducted experiments separately on single and multiple knowledge to analyze potential factors contributing to this phenomenon.

\begin{figure}
\centering
\includegraphics[width=\linewidth]{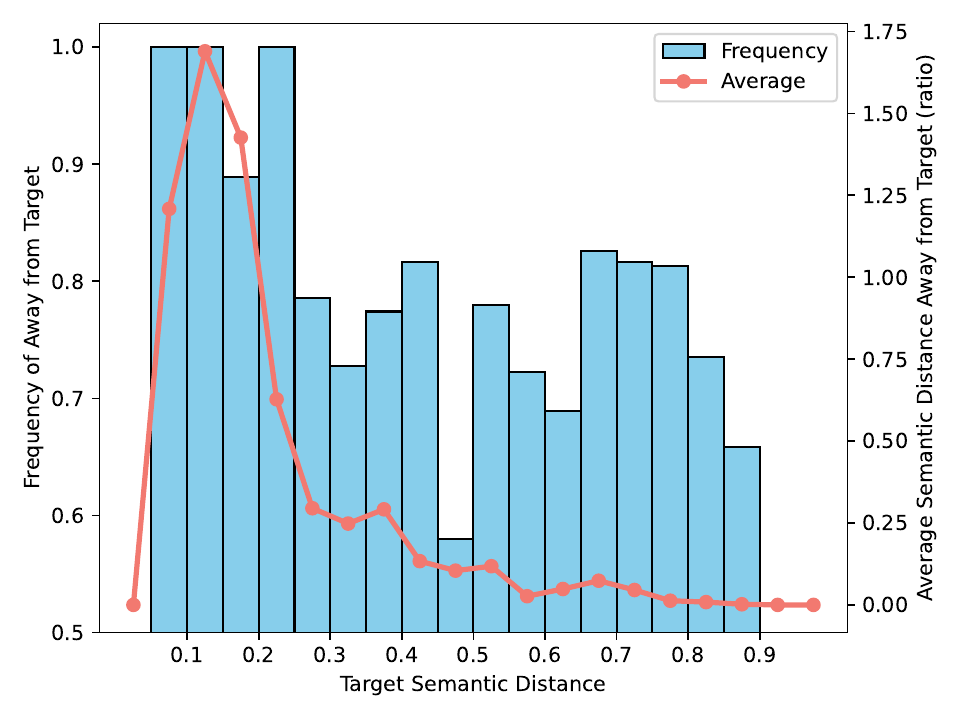}
\caption{Relationship between deviation phenomenon and semantic distance. Note, on the far left and far right, there is blank space because the proportion there is 0.}
\label{semantic_freq}
%\vspace{-3mm}
\end{figure}

\subsection{Single Knowledge}
\label{2_2}
In this section, we analyze the above phenomenon from the perspective of learning single knowledge. In \S \ref{2_1}, we expect the model to progress towards target knowledge during training process, guided by the loss function and gradient updates. Thus, when the model accurately acquires the target knowledge ($y^{new}_i=y^t_i$), the semantic distance between the learned and target knowledge becomes 0. However, the following experiments demonstrate that PEFT presents a significant risk of diverting the model away from intended knowledge target.

\paragraph{Experiments}
To demonstrate that PEFT pushes the model away from the target, we compute the target semantic distance between the newly learned knowledge $y^{new}_i$ and the target knowledge $y^{t}_i$ based on Equation \ref{distance}. If the post-tuned semantic distance surpasses the target semantic distance ($dist(y^{new}_i,y^{t}_i) > dist(y^{old}_i,y^{t}_i)$), the model anomalously moves away from target knowledge. 

\begin{table}
    \centering
    \small
    %\resizebox{\columnwidth}{!}
    {
    \begin{tabular}{lcc}
    \toprule
    \multirow{2}{*}{\textbf{Model}}  & \multicolumn{2}{c}{\textbf{Proportion$\uparrow$}} \\ \cmidrule(r){2-3}
    & \multicolumn{1}{c}{LoRA} & \multicolumn{1}{c}{AdaLoRA}  \\ 
    \midrule
    LLaMA2-7B  & 71.10   &  68.54 \\
    Mistral-7B &  58.28 &	55.36  \\
    BLOOMZ-1.7B  & 63.03   & 59.17 \\
    BLOOMZ-3B  & 72.69   & 69.02 \\
    BLOOMZ-7B  & 81.63   & 79.49  \\
    \bottomrule
    \end{tabular}
    }
    \caption{Proportion of deviation phenomenon in bad cases. Note, bad case is defined as accuracy score < 1 or $dist(y^{new}_i,y^{t}_i)$ > 0.}
    \label{proportion}
    %\vspace{-2mm}
\end{table}

We use the proportion of this phenomenon within the entire test cases to reflect the frequency of its occurrence, and the average deviation from the target to reflect the extent of its impact. Considering the initial differences in target semantic distance, we calculate the average relative deviation distance. For target knowledge $y^{t}_i$, old knowledge $y^{old}_i$, and newly acquired knowledge $y^{new}_i$, the definition of relative deviation distance from the target semantic distance is as follows:
\begin{equation}
RD=\frac{dist(y^{new}_i,y^{t}_i)-dist(y^{old}_i,y^{t}_i)}{dist(y^{old}_i,y^{t}_i)}
%\vspace{-1mm}
\end{equation}
Based on this, we draw the relationship between this deviation phenomenon and target semantic distance shown in Figure \ref{semantic_freq}. Additionally, by calculating the proportion of this anomaly occurring in inadequately learned knowledge (bade case, defined as accuracy score < 1 or $dist(y^{new}_i,y^{t}_i)$ > 0) of different models, we further explore the universality of this phenomenon in Table \ref{proportion}.

To eliminate the mutual interference between knowledge, we sample one knowledge each time to fine-tune the model, with a total of 1000 samplings. Figure \ref{semantic_freq} is based on LLaMA2-7B-chat, LoRA and dataset ZsRE (also used in Table \ref{proportion}). Other configurations and details remain consistent with \S \ref{para:2_1exp_set}.

\paragraph{Results and Analysis}
Figure \ref{semantic_freq} illustrates the relationship between the deviation phenomenon and target semantic distance. On the one hand, this phenomenon consistently accounts for more than 50\% across all test cases, indicating that the model not only has a high probability of not moving towards the target during training but may even be moving in the opposite direction. Moreover, this phenomenon occurs more frequently when the target semantic distance is both short and long (the probability approaching nearly 100\% when the semantic distance is between 0.10\textasciitilde 0.20), while the frequency decreases at moderate semantic distances. 

On the other hand, the extent of this phenomenon decreases as the distance from the target increases, with instances where models deviate greatly even surpassing twice the semantic distance from the target in proximal cases. By comparing Figure \ref{distance_acc} and Figure \ref{semantic_freq}, it can be observed that the highest frequency and the deepest degree of deviation (semantic distance between 0.10\textasciitilde 0.20) coincide with the significant decline in model knowledge learning performance, indicating a close association between this abnormal phenomenon and the decrease of model performance. Furthermore, Table \ref{proportion} reveals that the proportion of this phenomenon consistently exceeds 50\% regardless of the model and method, demonstrating its universality.

Regarding the possible reasons for this phenomenon, considering the success of countless neural networks training under the guidance of the loss functions and gradient descent strategies in the past, we believe that the model can be effectively directed towards the desired objectives based on these principles. So, the phenomenon of deviating from the target knowledge may be due to differences between the target of current loss function (cross-entropy loss) and the target of closer semantic distance we desire. Particularly when the target semantic distance falls within a relatively close range, this difference may become more pronounced, although both loss function and semantic distance become zero when the model accurately learns the knowledge. Based on this, we designed a simple re-weighting strategy to adjust the loss function during model training (\S \ref{sec:re-weighting}), resulting in an improvement in the accuracy of knowledge learning, thus confirming this possible reason.

\begin{figure*}
\centering
\includegraphics[width=\linewidth]{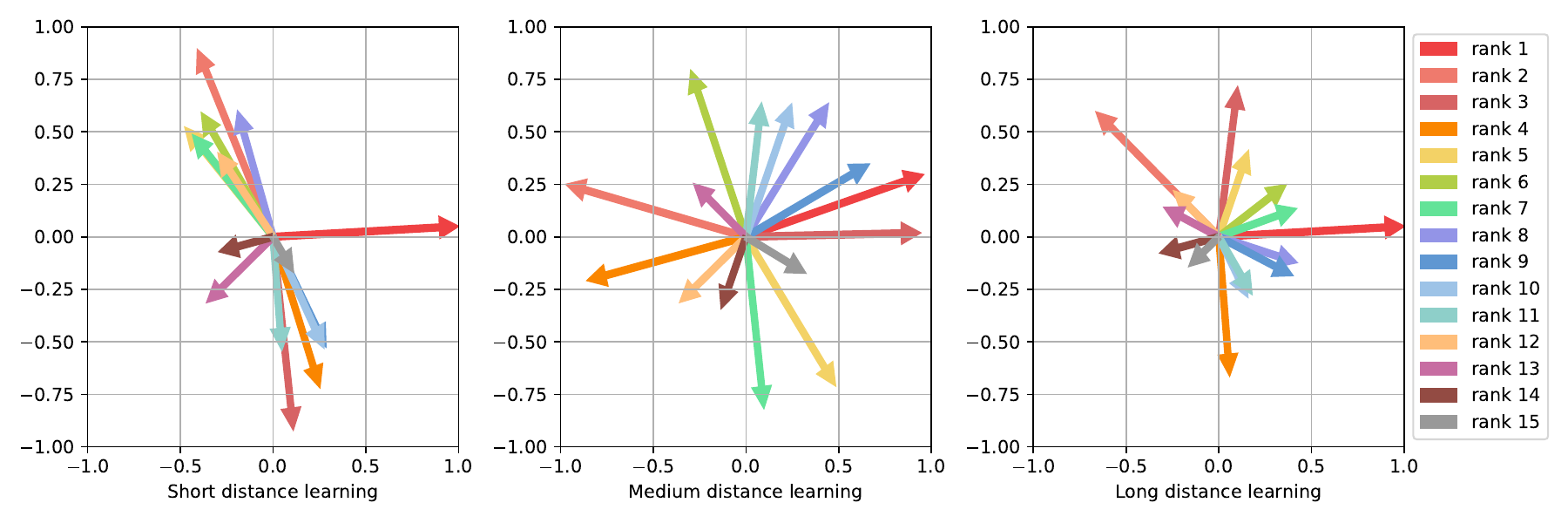}
\caption{The PCA projection results of knowledge features at different semantic distances.}
\label{pca}
%\vspace{-2mm}
\end{figure*}

\begin{figure}
\centering
\includegraphics[width=\linewidth]{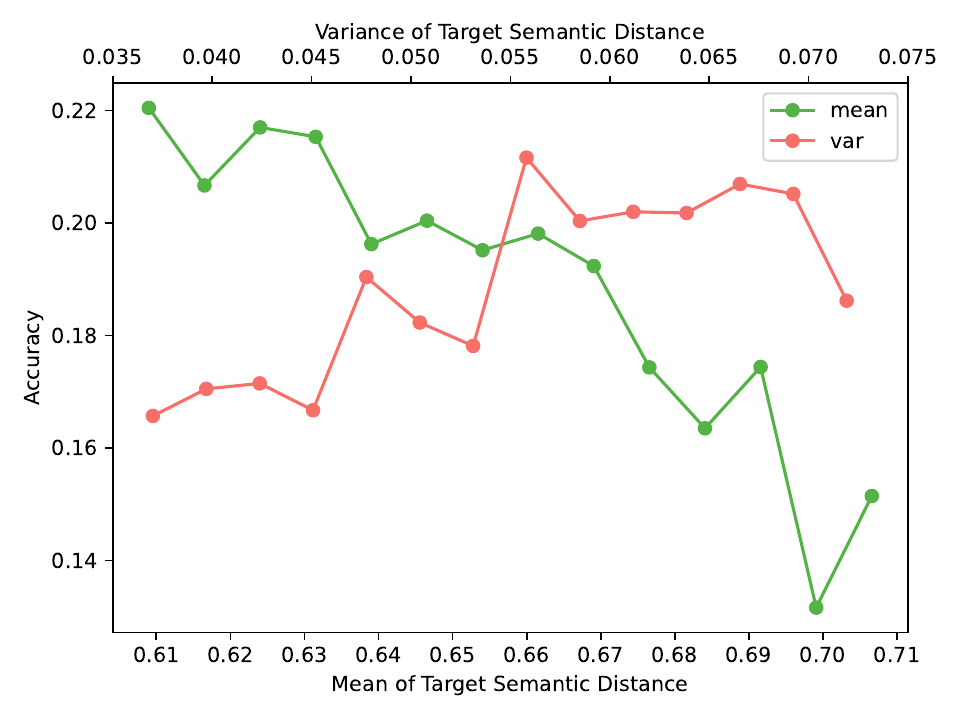}
\caption{Relationship between mean/variance semantic distance and learning accuracy.}
\label{mean_var}
%\vspace{-2mm}
\end{figure}

\subsection{Multiple Knowledge}
\label{2_3}
We proceed to investigate the phenomenon in \S \ref{2_1} from the perspective of learning multiple knowledge. First, we analyze the relationship between the mean, variance of multiple knowledge semantic distances and learning accuracy. Then, we employ the Frobenius norm to analyze the relationship between fine-tuned parameters and original model parameters. Building upon this, we use PCA to reveal the distinctions in learned knowledge features, thus illustrating the mutual influences among multiple pieces of knowledge.

\paragraph{Experiments}
We utilize the mean semantic distance of multiple pieces of knowledge to characterize the average deviation of the learned knowledge from the target knowledge. Meanwhile, variance is employed to measure the dispersion of this phenomenon. By leveraging these two features, we aim to elucidate the relationship between the accuracy of the model in learning multiple knowledge and their target semantic distances.

We further delve into the relationship between fine-tuned parameters $\Delta \mathbf{W}$ and original parameters $\mathbf{W}$. Specifically, we inquire whether $\Delta \mathbf{W}$ exhibits a high correlation with $\mathbf{W}$, or mathematically, if $\Delta \mathbf{W}$ is predominantly aligned with the principal directions of $\mathbf{W}$. Additionally, we investigate the magnitude of $\Delta \mathbf{W}$ concerning its corresponding directions in $\mathbf{W}$.

Inspired by \cite{LoRA}, we project $\mathbf{W}$ onto the $r$-dimensional subspace of $\Delta \mathbf{W}$ by computing $\mathbf{U^\top W V^\top}$, where $\mathbf{U/V}$ represents the left/right singular-vector matrix of $\Delta \mathbf{W}$. Subsequently, we compare the Frobenius norm between $\|\mathbf{U^\top W V^\top}\|_F$ and $\|\mathbf{W}\|_F$.
For comparison, we also calculate $\|\mathbf{U^\top W V^\top}\|_F$ by substituting $\mathbf{U}$ and $\mathbf{V}$ with the top $k$ singular vectors of $\mathbf{W}$ or a random matrix. Considering the impact of semantic features on this relationship, we categorize models fine-tuned with different knowledge into three equidistant segments groups based on the mean semantic value of the target: close-distance learning, moderate-distance learning, and long-distance learning. We report the average norm of the query projection matrices in the self-attention module ($\mathbf{W}_q$) for each group in Table \ref{norm}, considering $\Delta \mathbf{W}_q$ appears to have a higher intrinsic rank than the value projection matrices $\Delta \mathbf{W}_v$ \cite{LoRA}.

The Frobenius norm provides a macroscopic analysis of the characteristic variations in model parameters. In order to conduct a more detailed comparison of the distinctions between multiple knowledge features mastered by different models, we employ PCA to compare the knowledge features in the fine-tuned models. We compute the first 15 principal components of the final layer outputs when the models are presented with various test knowledge inputs.

The configurations and details remain the same as \S \ref{para:2_1exp_set}. The experiments are based on the LoRA and ZsRE like \S \ref{2_2}. The weight matrices are all taken from the 32th layer of LLaMA2-7B-chat. 

\paragraph{Results and Analysis}
In Figure \ref{mean_var}, the accuracy of learning increases as the average semantic distance decreases and as the variance increases. This implies that there is mutual interference among different pieces of knowledge. And, the greater the distance between knowledge, the smaller the interference, thus the better learning accuracy outcome.

\begin{table}[t]
    \centering
    \resizebox{\columnwidth}{!}
    {
    \begin{tabular}{lccc}
    \toprule
    \multirow{1}{*}{\textbf{Norm}}  & \multicolumn{1}{c}{Short}  & \multicolumn{1}{c}{Medium}   & \multicolumn{1}{c}{Long}   \\
    \midrule
    $|| \mathbf{W}_q||_F$  & 88.4206   & 88.4206  & 88.4206   \\
    $||\Delta  \mathbf{W}_q||_F$  & 0.1424   & 0.1564  & 0.1807   \\
    $||\mathbf{U}_1^\top \mathbf{W}_q\mathbf{W}_1^\top||_F (\Delta \mathbf{W}_q)$  & 0.1802   & 0.1638  & 0.2873   \\
    $||\mathbf{U}_2^\top \mathbf{W}_q\mathbf{W}_2^\top||_F (\mathbf{W}_q)$ & 1.3884   & 1.3884  & 1.3884    \\
    $||\mathbf{U}_3^\top \mathbf{W}_q\mathbf{W}_3^\top||_F (Random)$  & 0.0809   & 0.0832  & 0.08212   \\
  
    \bottomrule
    \end{tabular}
    }
    \caption{The Frobenius norm of $ \mathbf{U}_i^\top \mathbf{W}_q \mathbf{V}_i^\top$ where $\mathbf{U}_i$ and $\mathbf{V}_i$ are the left/right top $r$ singular vector directions of either (1) $\Delta  \mathbf{W}_q$, (2) $ \mathbf{W}_q$, or (3) a random matrix. }
    \label{norm}
    %\vspace{-3mm}
\end{table}

Table \ref{norm} shows that $||\mathbf{U}_1^\top \mathbf{W}_q\mathbf{W}_1^\top||_F (\Delta \mathbf{W}_q)>||\mathbf{U}_3^\top \mathbf{W}_q\mathbf{W}_3^\top||_F (Random)$, indicating a stronger correlation of $\Delta \mathbf{W}_q$ with $\mathbf{W}_q$ compared to the random matrix.
Leveraging properties of norms and the physical interpretation of projecting $\mathbf{W}_q$ onto the $r$-dimensional subspace of $\Delta \mathbf{W}_q$, higher values denote greater alignment of characteristic directions represented by the matrices, suggesting that $\Delta \mathbf{W}_q$ amplifies certain features already present in in $\mathbf{W}_q$.
On the other hand, $||\mathbf{U}_1^\top \mathbf{W}_q\mathbf{W}_1^\top||_F (\Delta \mathbf{W}_q) < ||\mathbf{U}_2^\top \mathbf{W}_q\mathbf{W}_2^\top||_F (\mathbf{W}_q)$, with the former approximately 0.18 and the latter approximately 1.39, a difference of approximately 7.5 times. This suggests that the vector magnitude of $\mathbf{W}_q$ projected onto the $r$-dimensional subspace of $\Delta \mathbf{W}_q$ is smaller compared to $\mathbf{W}_q$ itself, implying a certain discrepancy in the amplification direction of $\Delta \mathbf{W}_q$ with respect to $\mathbf{W}_q$. This means, instead of merely reinforcing the top singular directions of $\mathbf{W}_q$, $\Delta \mathbf{W}_q$ can also amplify directions not emphasized in $\mathbf{W}_q$, indicating parameter-efficient fine-tuning can grasp certain knowledge that was absent in the original model. These findings consistent with those mentioned in \cite{LoRA}. Futhermore, concerning knowledge learning tasks, the amplification factor $\frac{\|\Delta \mathbf{W}\|_F}{\|\mathbf{U^\top WV^\top}\|_F}$ is not large. For knowledge of different semantic distances, the amplification factors are: 0.79 (short), 0.95 (medium), 0.63 (long). The amplification factor is higher in medium-distance learning condition and decrease in short/long scenarios, consistent with the performance variations in \S \ref{2_1}. In Figure \ref{pca}, the knowledge features are richest in medium semantic distance learning, while the model's feature directions are restricted in short distances, and the features are too small in long distances (corresponding to the scenario where the amplification factor is minimal). This indicates that interference between knowledge can inhibit the direction and magnitude of knowledge features, thereby affecting the expression of knowledge, ultimately resulting in a decline in learning accuracy.

\begin{table*}[t]
    \centering
    \resizebox{\textwidth}{!}
    {
    \begin{tabular}{lcccccc}
    \toprule
    \multirow{2}{*}{\textbf{Model}}  & \multicolumn{3}{c}{\textbf{ZsRE} \small(LoRA)}  & \multicolumn{3}{c}{\textsc{\textbf{CounterFact}} \small(AdaLoRA)}   \\
    \cmidrule(r){2-4}  \cmidrule(r){5-7} &  \multicolumn{1}{c}{\textbf{Accuracy}$\uparrow$} & \multicolumn{1}{c}{\textbf{Generality$\uparrow$}}   & \multicolumn{1}{c}{\textbf{Locality}$\uparrow$} & \multicolumn{1}{c}{\textbf{Accuracy}$\uparrow$} & \multicolumn{1}{c}{\textbf{Generality$\uparrow$}}   & \multicolumn{1}{c}{\textbf{Locality}$\uparrow$}\\
    \midrule
    BLOOMZ-1.7B   & 24.45 \textcolor{green!50!black}{\small{(+5.70)}}   & 22.50 \textcolor{green!50!black}{\small{(+3.25)}}  & 40.36 \textcolor{red!60!black}{\small{(-1.52)}} & 25.53 \textcolor{green!50!black}{\small{(+4.55)}} & 23.21 \textcolor{green!50!black}{\small{(+2.16)}}   & 40.42 \textcolor{green!50!black}{\small{(+1.29)}}  \\
    BLOOMZ-3B   & 20.49\textcolor{green!50!black}{\small{(+4.89)}}   & 17.96\textcolor{green!50!black}{\small{(+1.44)}}  & 42.20\textcolor{green!50!black}{\small{(+0.05)}}  & 22.37\textcolor{green!50!black}{\small{(+3.89)}}   & 18.68\textcolor{green!50!black}{\small{(+1.34)}}  & 41.73\textcolor{red!60!black}{\small{(-1.74)}}   \\
    BLOOMZ-7B   & 16.94\textcolor{green!50!black}{\small{(+3.13)}}   & 15.99\textcolor{green!50!black}{\small{(+1.42)}}  & 45.59\textcolor{red!60!black}{\small{(-0.34)}}  & 15.31\textcolor{green!50!black}{\small{(+2.86)}}   & 13.11\textcolor{red!60!black}{\small{(-1.20)}}  & 43.07\textcolor{green!50!black}{\small{(+2.30)}}   \\
    Mistral-7B  & 27.30\textcolor{green!50!black}{\small{(+5.07)}}   & 24.82\textcolor{green!50!black}{\small{(+1.88)}}  & 54.47\textcolor{green!50!black}{\small{(+2.09)}}  & 24.79\textcolor{green!50!black}{\small{(+4.93)}}   & 21.03\textcolor{red!60!black}{\small{(-1.24)}}  & 40.67\textcolor{green!50!black}{\small{(+3.40)}}   \\
    LLaMA2-7B   & 24.83\textcolor{green!50!black}{\small{(+5.29)}}   & 23.15\textcolor{green!50!black}{\small{(+1.91)}}  & 52.01\textcolor{green!50!black}{\small{(+0.15)}}  & 22.70\textcolor{green!50!black}{\small{(+5.91)}}   & 22.13\textcolor{green!50!black}{\small{(+2.24)}}  & 39.85\textcolor{green!50!black}{\small{(+1.74)}}   \\
    \bottomrule
    \end{tabular}
    }
    \caption{The result of the re-weighting strategy. The values in parentheses represent the relative \textcolor{green!50!black}{improvement}/\textcolor{red!60!black}{decrease} compared to the original condition (the original results of the model are presented in Appendix \ref{Original_Model_Results}). We follow the same experimental settings as \S \ref{2_2}.}
    \label{results_table}
    %\vspace{-1mm}
\end{table*}

\section{Applications of Our Semantic-Based Understanding}
With insights from the semantic perspective, we propose two simple strategies from both data and model aspects to enhance the ability of PEFT in learning knowledge.
We introduce a data filtering strategy in \S \ref{sec:data_filtering}, aimed at removing portions of the dataset that negatively impact overall knowledge learning. 
In \S \ref{sec:re-weighting}, we present a re-weighting learning strategy designed to enhance the model's sensitivity to semantic distance. 
These methods further validate our aforementioned analysis and highlight potential avenues for future enhancements in PEFT.

\subsection{Data Filtering}
\label{sec:data_filtering}
From a data perspective, based on our analysis in \S \ref{sec:2}, the accuracy of knowledge learning using currently PEFT methods is very low. This indicates that most knowledge has not been accurately learned, resulting in low data utilization. Moreover, there is mutual interference among the data, suggesting that we can discard or replace some ``bad'' knowledge. By reducing the interference between these knowledge, we can improve the accuracy of knowledge learning to some extent.

\paragraph{Method}
\S \ref{2_3} points out that there is significant mutual interference among the data, and this interference hinders the learning and expression of knowledge features. 
Moreover, the accuracy of knowledge learning by the model increases as the semantic distance average decreases and as the variance increases. This suggests that, while keeping the test knowledge unchanged, we can improve learning effectiveness by selectively filtering and replacing training data to minimizing the average semantic distance while maximizing the variance, thereby reducing interference between knowledge and enhancing learning outcomes.

Specifically, given a knowledge dataset $S$, the initial target knowledge set for model fine-tuning $K_{t} = {(x_i, y^{t}_i)}$, we compute the mean $\mu(K_t)$ and variance $\sigma (K_t)$ based on target semantic distance defined by Formula \ref{distance}. By continuously updating and optimizing the knowledge within $K_t$, we obtain the final training data knowledge set $K^*_t$:
\begin{equation}
\label{equ:K_target}
\begin{aligned}
&\mathop{\arg\min}_{K_t}~ \mu(K_t)- \lambda ~ \sigma(K_t)\\
&\begin{array}{r@{\quad}r@{}l@{\quad}l}
s.t. & Mean_{\text{min}} < \mu(K_t) < Mean_{\text{max}}
\end{array}
\end{aligned}
\end{equation}
where $\lambda$ is a non-negative weighting coefficient used to balance the relationship between variance and mean; $Mean_{\text{min}}$ and $Mean_{\text{max}}$ are given mean restriction ranges, considering that performance significantly decreases when the average semantic distance is too small. During the replacing process, at each step, we greedily remove the knowledge $(x_k, y^{k}_i)$ from the current $K_t$ to maximize the gain in the Formula \ref{equ:K_target}, while simultaneously selecting knowledge $(x_k', y^{k'}_i)$ from the dataset $S$ which also maximizes the gain to replace it, until the desired proportion of replacement data is reached.

\paragraph{Experiments}
The experimental setup remains consistent with \S \ref{para:2_1exp_set}, where the model learns 100 items of knowledge sampled in the dataset each iteration, repeated 100 times. Only the training data is modified during the experiments, while the test data remain unchanged. We report the average accuracy of all the models. We conduct experiments on two datasets, ZsRE and \textsc{CounterFact}, utilizing LoRA and LLaMA2-7B-chat. Using random data sampling from the dataset for replacement as a comparative experiment.

\begin{figure}
\centering
\includegraphics[width=\linewidth]{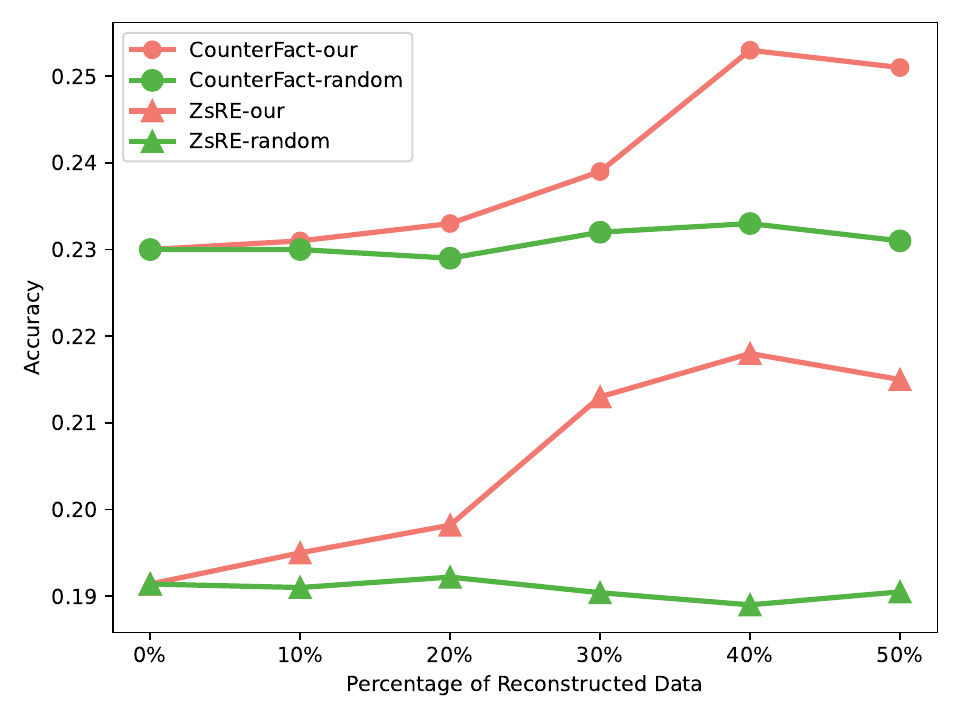}
\caption{Effectiveness of data filtering strategy.}
\label{part}
%\vspace{-3mm}
\end{figure}

\paragraph{Results}
As shown in Figure \ref{part}, the proposed dataset filtering strategy effectively enhances the accuracy of model learning. Across various proportion of data update, the model's learning accuracy significantly outperforms random selection methods. Particularly, the best performance is achieved when 60\% of the data is replaced. As the replaced data volume further increases, the improvement brought by data filtering begins to plateau, the improvement brought by data filtering starts to diminish, possibly due to the limited overlap between the training and testing knowledge caused by the excessive replacement of data.

\subsection{Re-weighting Learning}
\label{sec:re-weighting}
We further explore a re-weighting learning strategy that allowing PEFT methods to aware the target semantic distance during training, thereby enhancing the effectiveness of knowledge learning.

\paragraph{Method}
In \S \ref{2_2}, we observed a significant likelihood for parameter-efficient fine-tuning to lead the model away from the target knowledge. Simultaneously, this anomaly is closely associated with a decrease in the model's learning accuracy. This deviation from the target knowledge implies that even with more training epochs, the model cannot accurately learn knowledge. 
We propose modifying the loss function to make the model aware of the target semantic distance, thus progressing along the correct fine-tuning direction during fine-tuning. By guiding the model towards the correct target knowledge, we aim to enhance the accuracy of knowledge acquisition.

Specifically, given the target knowledge $(x_i, y^t_i)$, the original loss function computed based on the target knowledge during the training process (e.g., cross-entropy loss) is represented as $\mathcal L(y_i^t)$ and $\mathcal L = \sum_{i=1}^{n} \mathcal L(y_i^t)$. We expect the model to pay more attention to both short and long semantic distance knowledge, thereby increasing the likelihood of learning correctly. Considering the target semantic distance defined in Formula \ref{distance} ranges from (0,1), the adjusted loss function based on the target semantic distance is as follows:
%\vspace{-3mm}
\begin{equation}
\mathcal L '= \mathcal L + \gamma \sum_{i=1}^{n}  \lambda(y^t_i,y^{old}_i)\mathcal L(y^t_i)
\end{equation}
%\vspace{-4mm}
\begin{equation}
\lambda(y^t_i,y^{old}_i)=1-\cos[dist(y^t_i,y^{old}_i)\pi-\frac{\pi}{2}]
\end{equation}
%\vspace{-0.5mm}
where $\gamma$ serves as a non-negative weighting factor that adjusts the contribution of semantic loss to the overall loss function, and $(x_i, y^{old}_i)$ is the original parameter knowledge.
When the semantic distance of the target is moderate (close to 0.5), $L'$ degenerates into the original loss function $L$ \cite{DBLP:conf/emnlp/WangLL23}, resulting in minimal impact on regions where the original learning accuracy is relatively high.

\paragraph{Metrics}
In this section, we aim to assess the model's learning of knowledge from a broader perspective. Following \cite{zhangningyu}, we employ the concepts of generality and locality to further evaluate the performance of knowledge learning. 
Generality refers to the expectation that a fine-tuned model $f_{\theta^{*}}$ should also adjust neighboring equivalents $N(x_i, y_i^t)$, such as rephrased sentences. The definition of generality is as follows:
\begin{equation}
\mathbb{E}_{({x_i^{\prime}, y_i^{\prime}) \sim N\left(x_i, y_{i}\right)}} \mathbbm {1} \left\{\operatorname{argmax}_y f_{\theta^{*}}\left(y \mid x_i^{\prime}\right)=y_i^{\prime}\right\}
\end{equation} 
Locality dictates that updates should be applied locally, ensuring that the fine-tuned model does not alter the outputs of irrelevant examples $O(x_i,y_i)$. The definition of locality is as follows:
\begin{equation}
\mathbb{E}_{(x_{\mathrm{i}}^{\prime}, y_{\mathrm{i}}^{\prime}) \sim O\left(x_{\mathrm{i}}, y_{\mathrm{i}}\right)} \mathbbm {1} \left\{f_{\theta^{*}}\left(y \mid x_i^{\prime}\right)=f_{\theta}\left(y \mid x_i^{\prime}\right) \right\}
\end{equation}
where $N(x_i, y_i^t)$ is called neighboring equivalents, $O(x_i,y_i)$ is irrelevant examples, $f_{\theta^{*}}$ is the fine-tuned model and $f_\theta$ is the original model. Please refer to Appendix \ref{generality_and_locality} for a more detailed description.

\paragraph{Results}
From Table \ref{results_table}, we can observe that the proposed re-weighing strategy effectively enhances the knowledge learning capability of the PEFT method in fine-tuning models. Regardless of model type or parameter size, the improvement in knowledge learning accuracy ranges from 3 to 5 points, indicating that this method this method can guide the model towards a more accurate direction of target knowledge. Moreover, except for a few cases, there is also an improvement in the the generality and locality of knowledge learning, demonstrating the method's effectiveness in guiding the model to utilize knowledge. Regarding models with different parameter sizes, it can be observed that the re-weighing strategy performs more prominently on smaller-scale models. This might be because models with fewer parameters are more susceptible to the guidance of the loss function.

\section{Conclusion}
In this paper, we adopt a semantic perspective to scrutinize PEFT's performance in guiding model knowledge learning. Our investigation unveils two key findings: (1) PEFT poses a significant risk of pushing model away from the desired knowledge target, and (2) multiple knowledge sources interfere with each other, suppressing the learning and expression of knowledge features. Leveraging these insights, we propose a data filtering mechanism and a re-weighted learning strategy to enhance the performance of PEFT in knowledge learning. Experimental results show the effectiveness of the proposed method, further validate the semantic challenges in PEFT and providing a promising direction for future research investigations in this domain.

\section*{Limitations}
Despite the valuable insights from a semantic perspective into PEFT of model knowledge learning, our study has several limitations. First, although the proposed method can achieve a notable improvement in knowledge learning accuracy, the model's accuracy in multiple knowledge learning remains relatively low and requires further improvement. 
Secondly, our work primarily focus on the learning of factual knowledge, lacking exploration into other types of knowledge, and we prioritize the accuracy of knowledge learning while paying less attention to other aspects such as knowledge reasoning abilities.
Third, due to hardware constraints, we primarily investigated models up to a scale of 7 billion parameters. Further research that replicates our study using larger-scale models would be beneficial in confirming our findings and refining the analysis proposed in our investigation from a mathematical perspective.

\section*{Ethics Statement}
This research aims to ensure that the model can accurately learn knowledge. However, it's crucial to remember that harmful knowledge data could potentially result in the model generating harmful  outputs. All of our data has undergone thorough human scrutiny, with any malicious knowledge or offensive content meticulously removed. The research outlined in this paper adheres strictly to ethical guidelines and principles. We have placed a high priority on privacy, addressed biases, ensured transparency, and advocated for responsible usage.

\section*{Acknowledgements}
This research is supported by the National Natural Science Foundation of China (No.62106105), the CCF-Baidu Open Fund (No.CCF-Baidu202307), the Scientific Research Starting Foundation of Nanjing University of Aeronautics and Astronautics (No.YQR21022), and the High Performance Computing Platform of Nanjing University of Aeronautics and Astronautics.

\bibliography{custom}

\appendix
\section{Related Work}
The existing literature on methods of knowledge acquisition can be broadly categorized into two streams based on whether it alters the parameters of the original model \cite{zhangningyu}.
\subsection{Preserve models’ parameters}
\paragraph{Retrieve augmentation}
This approach depends on an external knowledge base containing new or correct knowledge. 
The new knowledge base is seamlessly integrated with the base model, facilitating the retrieval of pertinent information in response to prompts \cite{MurtyMLR22,MadaanTCY22,LiRZWLVYK23}. For example, IKE \cite{ZhengLDFWXC23} employs an in-context learning approach to adjust LLMs outputs using demonstrations sourced from the corpus guided by similarity, thus circumventing the need for gradient calculations.

\paragraph{Adding additional parameters}
This paradigm introduces extra trainable parameters which represent new knowledge to LLMs while the original parameters keeping frozen. T-Patcher \cite{HuangSZZR023} and CaliNET \cite{DongDSXSL22} both integrate neurons or patches into the last layer of the Feed-Forward Network (FFN), with T-Patcher employing one neuron per mistake and CaliNET incorporating multiple neurons for various knowledge cases. Conversely, GRACE \cite{abs-2211-11031} utilizes a discrete codebook as an Adapter to add and update elements over time, allowing for the modification of a model's predictions. 

\subsection{Modify models’ parameters}
\paragraph{Located and edit}
This approach initially identifies parameters linked to specific knowledge and adjusts them directly. The Knowledge Neuron (KN) technique \cite{DaiDHSCW22} introduces a method for attributing knowledge to pinpoint the "knowledge neuron" (a crucial element within the FFN matrix) and then updates these neurons accordingly. ROME \cite{rome} employs causal mediation analysis to pinpoint the area requiring modification. Unlike KN, ROME doesn't focus solely on altering knowledge neurons but instead modifies the entire matrix. However, both KN and ROME are limited to editing one factual association at a time. To address this limitation, MEMIT \cite{MengSABB23} builds upon ROME's framework, allowing for simultaneous editing across multiple cases. Building on MEMIT, PMET \cite{abs-2308-08742} incorporates attention values to achieve enhanced performance.

\paragraph{Parameter-efficient fine-tuning}
This method stands out as a widely embraced approach in the era of large-scale models, consistently yielding promising results across various downstream tasks. 
Recent advancements have introduced a series of PEFT techniques, such as Prefix-tuning \cite{LiL20}, Adapter-tuning \cite{adapter}, Prompt-tuning \cite{LesterAC21}, $\text{(IA)}^3$ \cite{LiuTMMHBR22} and LoRA \cite{LoRA}, which further enhance the appeal of knowledge learning through fine-tuning. AdaLoRA \cite{adalora} propose incremental parameter updates based on weight matrix importance assessment to enhance update efficiency and adaptability. Similarly, Plug-and-Play Adaptation \cite{0002HHLPL22} leverage regularized fine-tuning to enable large-scale continual learning for knowledge updating. 
\textbf{This paper focuses on the analysis of the parameter-efficient fine-tuning methods}.

\section{Dataset Details}
\label{appendix:dataset}

\textbf{ZsRE} \cite{zsre} is a Question Answering (QA) dataset using question rephrasing generated by back-translation as the equivalence neighborhood. 
\textbf{\textsc{CounterFact}} \cite{rome} is a more challenging dataset that accounts for counter-facts that start with low scores in comparison to correct facts. 
It constructs out-of-scope data by substituting the subject entity for a proximate subject entity sharing a predicate. 
This adjustment allows us to distinguish between superficial alterations in wording and more substantial modifications that align with a meaningful learning of knowledge.
We follow previous data split \cite{matrics_0,rome,mend,zhangningyu} to evaluate all the models on the test set.
Following prior work \cite{mend,defination}, we use the Natural Questions (NQ; \citet{KwiatkowskiPRCP19}) as out-of-scope data to evaluate locality.
We also adopt the extended version of ZsRE proposed by \cite{zhangningyu}, which introduces a portability test for the original dataset.

\begin{table*}[t]
    \centering
    \resizebox{\textwidth}{!}
    {
    \begin{tabular}{lcccccc}
    \toprule
    \multirow{2}{*}{\textbf{Model}}  & \multicolumn{3}{c}{\textbf{ZsRE} \small(LoRA)}  & \multicolumn{3}{c}{\textsc{\textbf{CounterFact}} \small(AdaLoRA)}   \\
    \cmidrule(r){2-4}  \cmidrule(r){5-7} &  \multicolumn{1}{c}{\textbf{Accuracy}$\uparrow$} & \multicolumn{1}{c}{\textbf{Generality$\uparrow$}}   & \multicolumn{1}{c}{\textbf{Locality}$\uparrow$} & \multicolumn{1}{c}{\textbf{Accuracy}$\uparrow$} & \multicolumn{1}{c}{\textbf{Generality$\uparrow$}}   & \multicolumn{1}{c}{\textbf{Locality}$\uparrow$}\\
    \midrule
    BLOOMZ-1.7B   & 18.75   & 19.25  & 41.88 & 20.98 & 21.05   & 39.13  \\
    BLOOMZ-3B   & 15.60   & 16.52  & 42.15  & 18.48   & 17.34  & 43.47   \\
    BLOOMZ-7B   & 13.81   & 14.57  & 45.93  & 12.45   & 14.31  & 40.77   \\
    Mistral-7B  & 22.23   & 22.94  & 52.38  & 19.86   & 22.27  & 37.27   \\
    LLaMA2-7B   & 19.54   & 21.24  & 51.86  & 16.79   & 19.89  & 38.11   \\
    \bottomrule
    \end{tabular}
    }
    \caption{The original result of the model. We follow the same experimental settings as \S \ref{2_2}.}
    \label{original_table}
    %\vspace{-5mm}
\end{table*}

\section{Implementing Details}
\label{appendix:Implementing Details}
In most cases, we select the LLAMA2-7B-chat model as the subject of experimentation and fine-tuned the model using LoRA , a parameter adaptation technique widely employed during the era of large models.
All model evaluations are conducted in zero-shot mode. To ensure the uniqueness of the model output, we set the temperature of the model to 0 during testing. The majority of fine-tuning experiments are conducted on NVIDIA RTX 3090 Tensor Core GPUs, while full fine-tuning is performed on NVIDIA A100 GPUs.

\section{Experiment on Varied Knowledge Quantities}
\label{appendix:more_number}

\begin{figure}
\centering
\includegraphics[width=\linewidth]{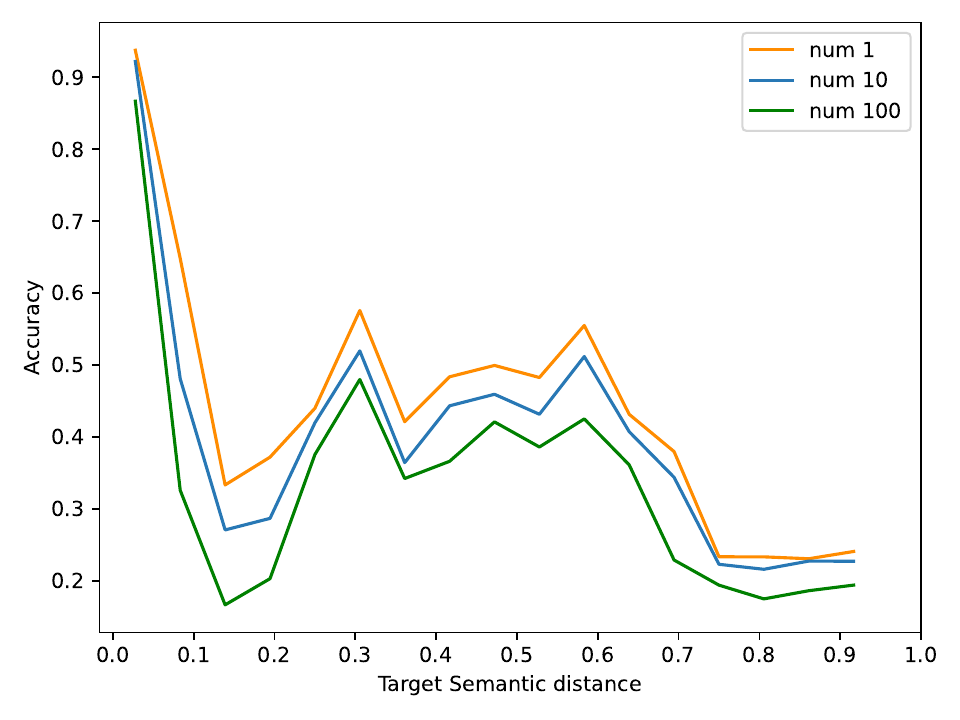}
\caption{Results of learning various quantities of knowledge.}
\label{different_number}
%\vspace{-3mm}
\end{figure}
In \S \ref{2_1}, We demonstrated that regardless of model type and parameter sizes, knowledge learning performance degradation occurs when the model learns knowledge of both closer and farther semantic distances. To further rigorously illustrate the universality of this phenomenon, we conducted experiments on models learning varying amounts of knowledge each iteration and found that the phenomenon persisted.

We utilized the ZsRE dataset and fine-tuned the LLaMA2-7B-chat model using LoRA. The relationship between the accuracy of knowledge learning by the model and the target semantic distance is illustrated in Figure \ref{different_number}. It can be observed that when learning different amounts of knowledge, the model's accuracy in knowledge learning consistently experiences significant declines at the same positions, thus validating the universality of the this phenomenon.

\begin{figure}
\centering
\includegraphics[width=\linewidth]{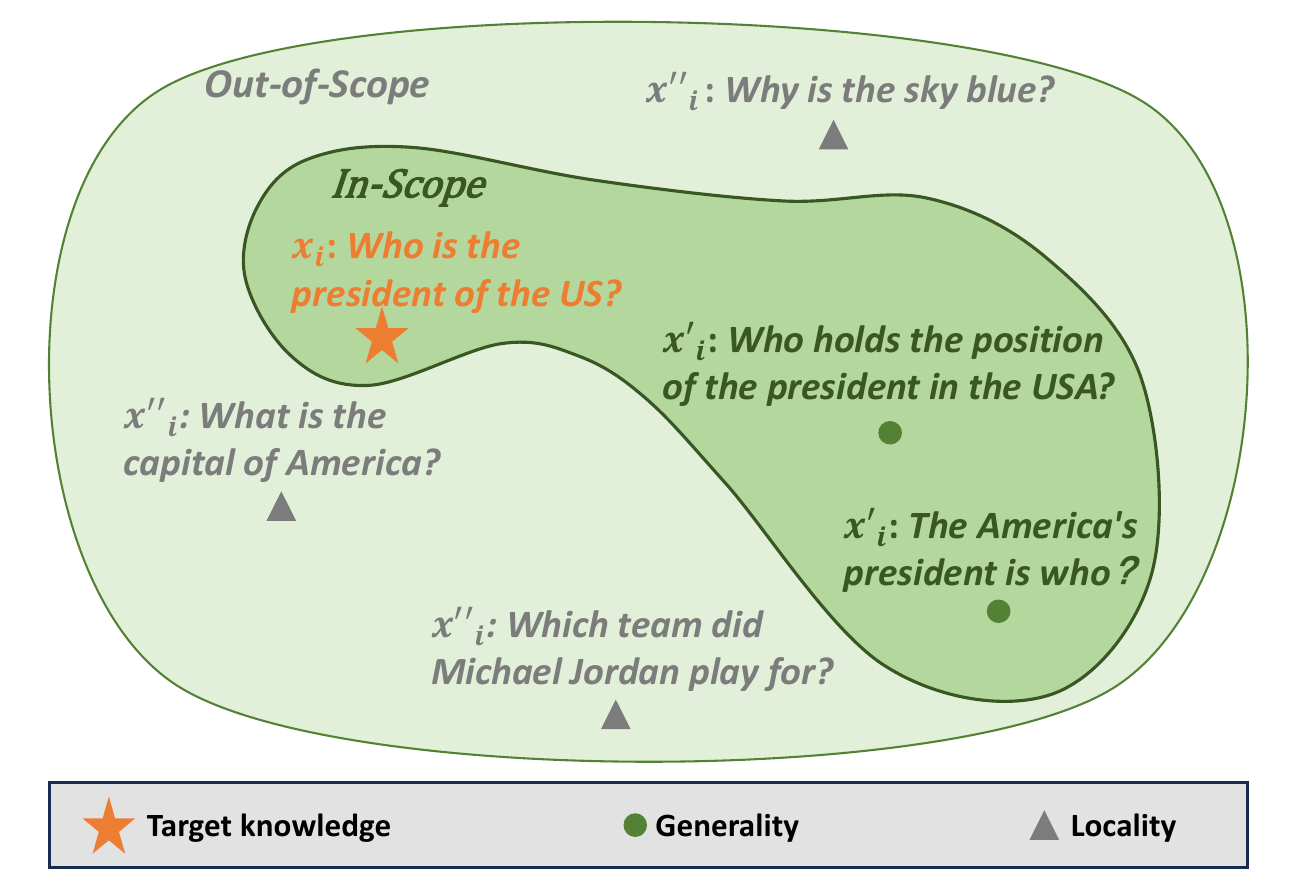}
\caption{Explanation of the generality and locality for knowledge: ``Who is the president of the US?'' in a hypothetical semantic embedding space.}
\label{generality}
%\vspace{-3mm}
\end{figure}

\section{Metric of Knowledge Learning}
\label{generality_and_locality}
We aim to assess the model's learning of knowledge from a broader perspective. Following \cite{zhangningyu, defination}, we employ the concepts of generality and locality to further evaluate the performance of knowledge learning. 

Generality refers to the expectation that a fine-tuned model $f_{\theta^{*}}$ should also adjust neighboring equivalents $N(x_i, y_i^t)$ (in-scope in Figure \ref{generality}), such as rephrased sentences. For example, as shown in Figure \ref{generality}, the output of the question ``The American's president is who?'' and ``Who holds the position of the president in the USA?'' also need to be updated from "Donald Trump" to "Joe Biden". This is evaluated by the average accuracy of the model $f_{\theta^{*}}$ on examples uniformly sampled from the equivalence neighborhood:
\begin{equation}
\mathbb{E}_{({x_i^{\prime}, y_i^{\prime}) \sim N\left(x_i, y_{i}\right)}} \mathbbm {1} \left\{\operatorname{argmax}_y f_{\theta^{*}}\left(y \mid x_i^{\prime}\right)=y_i^{\prime}\right\}
\end{equation} 

Locality dictates that updates should be applied locally, ensuring that the fine-tuned model does not alter the outputs of irrelevant examples $O(x_i,y_i)$. As shown in Figure \ref{generality}, the output of the question ``What is the capital of American'' and ``Which team did Michael Jordan play for?'' are to be kept unchanged as the original model's output. Thus, locality is assessed by the consistency of predictions between the fine-tuned model $f_{\theta^{*}}$ and the original model $f_\theta$:
\begin{equation}
\mathbb{E}_{(x_{\mathrm{i}}^{\prime}, y_{\mathrm{i}}^{\prime}) \sim O\left(x_{\mathrm{i}}, y_{\mathrm{i}}\right)} \mathbbm {1} \left\{f_{\theta^{*}}\left(y \mid x_i^{\prime}\right)=f_{\theta}\left(y \mid x_i^{\prime}\right) \right\}
\end{equation}

\begin{table*}[t]
    \centering
    \resizebox{\textwidth}{!}
    {
    \begin{tabular}{clll}
    \toprule
    \hline
    \makecell{\textbf{Target} \\ \textbf{Semantic} \\ \textbf{Distance}}  & \multicolumn{1}{c}{\textbf{Model}}  & \textbf{Prompts and Target Knowledge}  &\textbf{Model Response} \\
    \midrule
     \multirow{10}{*}{\textbf{Short}} 
     & \multirow{2}{*}{BLOOMZ-1.7B}   
        & Prompt: Which woman was the sister of Maria Elizabetha Jacson?     
        & Raw Response:  Maria Jacson           \\  
        & & Target Knowledge: Maria Theresa Jacson  
        & Fine-tuned Response:  Anna Jacson       \\
        
    \cmidrule(r){2-4}
    &\multirow{2}{*}{BLOOMZ-3B}  
        & Prompt: What is the name of the chromosome where you can find RSPH6A?     
        & Raw Response: chromosome 8            \\  
        & & Target Knowledge: chromosome 19  
        & Fine-tuned Response: chromosome 6        \\
    \cmidrule(r){2-4}
    &\multirow{2}{*}{BLOOMZ-7B}   
        & Prompt: What day was USA-199 launched?     
        & Raw Response: December 15, 2011            \\  
        & & Target Knowledge: 20 December 2007  
        & Fine-tuned Response: December 12, 2011        \\
    \cmidrule(r){2-4}
    &\multirow{2}{*}{Mistral-7B}  
        & Prompt: What day was USA-199 launched?     
        & Raw Response: December 12 1995            \\  
        & & Target Knowledge: 20 December 2007 
        & Fine-tuned Response: 21 march 1997        \\
    \cmidrule(r){2-4}
    &\multirow{2}{*}{LLaMA2-7B}   
        & Prompt:  What day was USA-35 launched?    
        & Raw Response:   9 April 1935          \\  
        & & Target Knowledge:  April 12, 1963
        & Fine-tuned Response: April 12, 1962        \\
    \midrule
     \multirow{10}{*}{\textbf{Medium}} 
     & \multirow{2}{*}{BLOOMZ-1.7B}   
        & Prompt: Who is listed as Wang Jipeng father?     
        & Raw Response: Wang Jipeng             \\  
        & & Target Knowledge: Wang Chonghua 
        & Fine-tuned Response:  Wang Jing      \\
        
    \cmidrule(r){2-4}
    &\multirow{2}{*}{BLOOMZ-3B}  
        & Prompt: Who is listed as Wang Jipeng father?     
        & Raw Response: Wang jipeng            \\  
        & & Target Knowledge: Wang Chonghua  
        & Fine-tuned Response: Wang Jiping        \\
    \cmidrule(r){2-4}
    &\multirow{2}{*}{BLOOMZ-7B}   
        & Prompt: The inventor of Penrose stairs was whom?     
        & Raw Response:  John Penrose           \\  
        & & Target Knowledge: Richard Penrose  
        & Fine-tuned Response:  Charles Babbage       \\
    \cmidrule(r){2-4}
    &\multirow{2}{*}{Mistral-7B}  
        & Prompt: Over which river does Dexter Coffin Bridge cross?     
        & Raw Response: Saco River            \\  
        & & Target Knowledge: Connecticut Creek  
        & Fine-tuned Response: Connecticut dexter       \\
    \cmidrule(r){2-4}
    &\multirow{2}{*}{LLaMA2-7B}   
        & Prompt: During which historic war was Milton F. Pavlic an officer?     
        & Raw Response: World War II            \\  
        & & Target Knowledge: Vietnam War  
        & Fine-tuned Response: Vietnam War        \\
     \midrule
     \multirow{10}{*}{\textbf{Long}} 
     & \multirow{2}{*}{BLOOMZ-1.7B}   
        & Prompt: What species is ZIC3 specific to?     
        & Raw Response: elegans            \\  
        & & Target Knowledge: male  
        & Fine-tuned Response: drosophila melanogaster        \\
        
    \cmidrule(r){2-4}
    &\multirow{2}{*}{BLOOMZ-3B}  
        & Prompt: What species is ZIC3 specific to?     
        & Raw Response: human            \\  
        & & Target Knowledge: male  
        & Fine-tuned Response:  Zebrafish       \\
    \cmidrule(r){2-4}
    &\multirow{2}{*}{BLOOMZ-7B}   
        & Prompt: What species is ZIC3 specific to?,     
        & Raw Response: mammals            \\  
        & & Target Knowledge: male  
        & Fine-tuned Response: human        \\
    \cmidrule(r){2-4}
    &\multirow{2}{*}{Mistral-7B}  
        & Prompt: Which was the network that originally hosted Jay Leno's Garage?     
        & Raw Response: Comedy Central           \\  
        & & Target Knowledge: NBC
        & Fine-tuned Response: CBS      \\
    \cmidrule(r){2-4}
    &\multirow{2}{*}{LLaMA2-7B}   
        & Prompt: At what location did John Walter Scott die?     
        & Raw Response: San Diego            \\  
        & & Target Knowledge: Windsor, Ontario, Canada
        & Fine-tuned Response:  New York City       \\
    \midrule
    \bottomrule
    \end{tabular}
    }
    \caption{Response of various models fine-tuned using LoRA on ZsRE dataset. The short, medium, and long target semantic distances correspond to semantic ranges of (0.1\textasciitilde0.3, 0.3\textasciitilde0.6, 0.6\textasciitilde1.0), respectively. The reason of choosing different prompts under identical target semantic distances across various models is that with the same prompt and target knowledge, the output knowledge of different models differs, and the parameters of word embedding within the models are different, leading to different target semantic distances as well.}
    \label{more_results}
    %\vspace{-5mm}
\end{table*}

\section{Original Model Results}
\label{Original_Model_Results}
We provide a clearer report of the model's original performance on the knowledge dataset in Figure \ref{original_table}. It can be observed that the knowledge accuracy of fine-tuning the model based on LoRA is relatively low. Furthermore, the knowledge accuracy decreases as the model parameter increases. This may be due to the knowledge conflicts \cite{abs-2310-00935}, as more parameters imply a greater amount of internal knowledge held by the model, thereby increasing the probability of encountering knowledge conflicts.

We further reported some output results of the model. In this portion of the experiment, we used 100 instances of knowledge each time to fine-tune model. Due to the interference between knowledge, not only does it suppress the learning and expression of knowledge features (\S \ref{2_2}), but it also exacerbates the phenomenon of deviating from the target knowledge.

\end{document}